%% file: main.tex
\documentclass{article} % For LaTeX2e
\usepackage{iclr2025_conference,times}

\input{math_commands.tex}

\usepackage{hyperref}
\usepackage{url}

\usepackage[utf8]{inputenc} % allow utf-8 input
\usepackage[T1]{fontenc}    % use 8-bit T1 fonts
\usepackage{hyperref}       % hyperlinks
\usepackage{url}            % simple URL typesetting
\usepackage{booktabs}       % professional-quality tables
\usepackage{amsfonts}       % blackboard math symbols
\usepackage{nicefrac}       % compact symbols for 1/2, etc.
\usepackage{microtype}      % microtypography
\usepackage{xcolor}         % colors

\usepackage{amssymb}
\usepackage{amsmath}
\usepackage{soul}
\usepackage{xcolor}
\usepackage{graphicx}
\usepackage{colortbl}   % Enables coloring in tables
\usepackage{environ}   % For handling environments
\usepackage{listings}
\usepackage{svg}

\newcommand{\cmark}{\textcolor{green}{\checkmark}}  % Green check mark
\newcommand{\xmark}{\textcolor{red}{\text{\sffamily x}}}  % Red X mark
\newcommand{\pmark}{\textcolor{blue}{$\triangle$}}
\newcommand{\comment}[1]{\sethlcolor{yellow}\hl{#1}}

\NewEnviron{hide_content}{}

\definecolor{SkyBlue}{rgb}{0.88, 0.92, 0.98}
\definecolor{PaleGreen}{rgb}{0.88, 0.98, 0.90}
\definecolor{LightYellow}{rgb}{0.98, 0.98, 0.88}

\newcommand{\ameet}[1]{\textcolor{red}{}}
\newcommand{\todo}[1]{\textcolor{red}{}}
\newcommand{\cd}[1]{\textcolor{magenta}{}}

\title{
The Impact of Element Ordering on LM Agent Performance
}

\author{Wayne Chi, Ameet Talwalkar \& Chris Donahue\\
School of Computer Science\\ 
Carnegie Mellon University\\
\texttt{\{waynechi, atalwalk, chrisdon\}@cs.cmu.edu}
}

\iclrfinalcopy % Uncomment for camera-ready version, but NOT for submission.
\begin{document}

\maketitle

% Footnote for code -> Uncomment for camera ready.
\begingroup
\renewcommand\thefootnote{}
\footnotetext{Our code is available at \url{https://github.com/waynchi/gui-agent}.}
\addtocounter{footnote}{-1}
\endgroup

\input{0_abstract}
\input{1_intro}
\input{2_related}
% \input{3_obs}
\input{4_method}

\input{4.1_feature_ablation}

\input{4.2_experimental_setup}
\input{5_results}
\input{6_future_work}
\input{7_conclusion}
\bibliographystyle{iclr2025_conference}
\bibliography{main}

% \bibliography{iclr2025_conference}
% \bibliographystyle{iclr2025_conference}

\appendix
\input{10_appendix}

\end{document}

%% file: math_commands.tex
\usepackage{amsmath,amsfonts,bm}

\def\eqref#1{equation~\ref{#1}}
\def\1{\bm{1}}

\DeclareMathAlphabet{\mathsfit}{\encodingdefault}{\sfdefault}{m}{sl}
\SetMathAlphabet{\mathsfit}{bold}{\encodingdefault}{\sfdefault}{bx}{n}

%% file: 0_abstract.tex
\begin{abstract}

There has been a surge of interest in language model agents that can navigate virtual environments such as the web or desktop.
To navigate such environments, agents benefit from information on the various elements (e.g.,~buttons, text, or images) present. 
However, it remains unclear 
which element attributes have the greatest impact on agent performance, 
especially in environments that only provide a graphical representation (i.e.,~pixels).
Here 
we find that the \emph{ordering} in which elements are presented to the language model is surprisingly impactful---randomizing 
element ordering in webpages compromises average agent performance to a degree comparable to removing all visible text from webpages. 
While web agents benefit from the semantic hierarchical ordering of elements available via the browser, agents that parse elements directly from pixels do not have access to any such ordering. 
Here we endeavor to derive effective orderings and investigate the impact of various element ordering methods in web and desktop environments. 
We find that dimensionality reduction provides a viable ordering for pixel-only environments.
We train a UI element detection model to derive elements from pixels and apply our findings to an agent benchmark---OmniACT---where we only have access to pixels.
Our method completes more than two times as many tasks on average relative to the previous state-of-the-art.

%an increasingly popular multimodal representation for virtual environments.
% While multimodal representations are beneficial, we find that agents still largely rely on a text representation.
% Surprisingly, element ordering is incredibly impactful; incorrect ordering alone degrades performance comparably to removing all visible text.
%Using these findings, we propose a method to navigating any desktop application given only the screenshot. 
%We find interactable elements using an object detection model, extract text through OCR, and order elements through dimensionality reduction.

\end{abstract}

%% file: 1_intro.tex
\section{Introduction}

There has been growing interest in using language model (LM) agents
%---also commonly known as Language Model Agents (LM Agents)---
to autonomously navigate virtual environments.
% \cd{Why the distinction between language agents and language model agents? Can we just call them language model agents and not have the parenthetical?}
% \comment{I just rarely see the term LM agents being tossed around in actual papers.}
% Two of the most popular agent tasks include Web agents \citep{web-arena, language-computer-tasks, zheng2024gpt4vision} and Code agents \citep{yang2024sweagent, jimenez2024swebench, OpenDevin}.
Autonomous web agents~\citep{web-arena, language-computer-tasks, zheng2024gpt4vision, gur2024realworld, he2024webvoyager} have become a particularly popular area of research. 
Typically, a web agent takes as input a task prompt from a user, observes a text and visual representation of the environment, and then outputs 
%a single or multiple 
one or more 
actions to execute the task in the environment.
Recently, research interest has expanded to include agents that can navigate mobile~\citep{rawles2023android, yan2023gpt4v} and desktop~\citep{xie2024osworld, kapoor2024omniact, bonatti2024windowsagentarenaevaluating} environments as well. 

% In either case, 
At a high level, a virtual environment consists of numerous elements---some are interactive (e.g. buttons or widgets), while others are not (e.g. plain text). 
To allow for human navigation, these elements are usually represented in the pixel space via a Graphical User Interface (GUI).
In contrast, 
%an agent must rely on some state representation to navigate a virtual environment.
agents often rely on distinct state representations to navigate virtual environments. 
The exact format of a state representation varies between environments and approaches.
In web environments, common text representations include the HTML or accessibility tree~\citep{web-arena, koh2024visualwebarena}.
For visual representations, a popular approach is to label UI elements with bounding boxes and numeric identifiers~\citep{koh2024visualwebarena, he2024webvoyager}, known as \emph{Set-of-Mark}~\citep{set-of-mark}.
In either case, the state representation is derived from the underlying Document Object Model (DOM)~\citep{web-arena, koh2024visualwebarena, he2024webvoyager}.
However, many environments lack a descriptive DOM and \textit{only} provide pixel information, which we show is insufficient for existing agents (see Section~\ref{sec:text-needed}).
% Later on, we show that existing agents still perform better when provided with text or Set-of-Mark as compared to a pixels only.
To construct an effective state representation from only pixels, it is important to answer the following basic questions about these representations: 
What aspects of a state representation are most important to an agent?
How can we derive these important aspects with only pixels?

\begin{figure}[!t]
    \centering
    % \includesvg[width=\textwidth]{imgs/fig1.svg}
    \includegraphics[width=\textwidth]{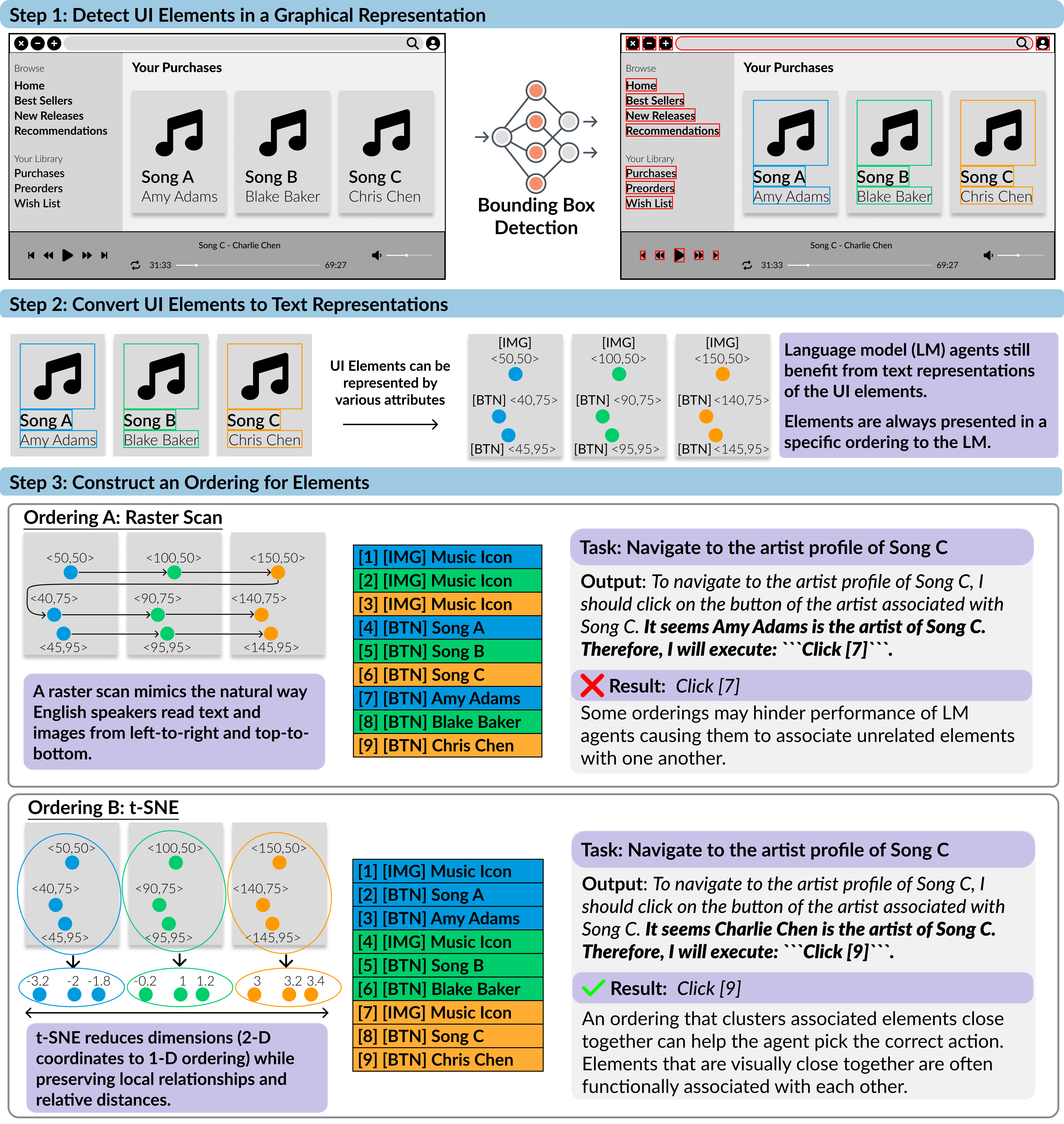}
    \caption{We are motivated by the goal of enabling agents to act on environments where an underlying DOM does not exist. Instead, the agent must determine its next action using only the environment's graphical representations.
    % We are motivated by the goal of enabling agents to act on environments that lack access to the DOM, instead relying purely on graphical representations as input.
    In Step 1, we first detect a list of unordered UI elements using an object detection model and identify them with bounding boxes.
    In Step 2, we convert these UI elements to their text representation.
    In Step 3, we order the elements via 2D-to-1D dimensionality reduction. 
    Due to the sequential nature of a language model, elements are always presented in a specific order to the LM Agent. 
    Finding an effective ordering is non-trivial, yet can significantly affect agent performance. Elements that are visually close together are often functionally associated with each other. t-SNE's ability to retain local structure allows it to generate an effective ordering.}
    %\ameet{Can you add more details to this caption to walk the reader through each of the 5 steps included in the visual?} \cd{+1, need more details in the caption here. Expecting something like ``Motivated by the goal of enabling agents to operate on raw pixels, \ldots''}}
    \label{fig:enter-label}
\end{figure}

In almost all implementations of the agent's state representation, there exists a list of interactable or non-interactable elements which the agent uses to determine the next action~\citep{koh2024visualwebarena, web-arena, he2024webvoyager, GPT4VAct, yan2023gpt4v, vimGPT, kapoor2024omniact, xie2024osworld}.
Elements are characterized by various attributes such as visual appearance, text descriptions, or type labels.
%Whether considered explicitly or implicitly, 
Because the state representation is the input to an LM, 
this list of elements is always 
%defined by 
presented in 
a specific ordering.
For example, the default method to derive elements from a webpage performs a pre-order traversal of the DOM tree~\citep{w3c2013selectors}.
We analyze various attributes of a popular state representation for agents and find element ordering to be the single most impactful attribute to agent performance.
We find that the ordering of elements can dramatically affect the performance of an agent, resulting in differences of up to 49\% relative performance.

\begin{hide_content}
At a high level, a virtual environment consists of numerous elements---some are interactive (e.g. buttons or widgets), while others are not (e.g. plain text). 
To allow for human navigation, these elements are usually represented in the pixel space via a Graphical User Interface (GUI).
While sometimes , environments provide \textit{only} pixel-level information. \ameet{this sentence is confusing to me, as  what else might be provided other than pixel-level info?  are you talking about DOM / accessibility tree?  if so, this sentence still feels out of place, since in the next paragraph you talk about text vs visual representations.}
To construct an effective state representation in these scenarios, it is crucial to answer the following questions.
What aspects of a state representation are most important to an agent?
How can we derive these important aspects with only pixel-level information?

The exact format of a state representation varies between environments and approaches.
Common choices for a text representation include the HTML or accessibility tree~\citep{web-arena, koh2024visualwebarena, xie2024osworld}.
For visual representations, a popular approach is to label UI elements with bounding boxes and numeric identifiers~\citep{set-of-mark, koh2024visualwebarena, he2024webvoyager}.
In almost all implementations, the agent observes a list of interactable or non-interactable elements to determine the next action.
Elements are characterized by various attributes such as visual appearance, text descriptions, or type labels.
%Whether considered explicitly or implicitly, 
Because the state representation is the input to an LM, 
this list of elements is always 
%defined by 
presented in 
a specific ordering.
For example, the default method to derive elements from a webpage performs a pre-order traversal of the Document Object Model (DOM) tree~\citep{w3c2013selectors}.
We analyze various attributes of a popular state representation for agents and find element ordering to be the single most impactful attribute.
\todo{something like ``where impact is defined as the reduction in performance when the ground truth information is removed''}. 
We find that the ordering of elements can dramatically affect the performance of an agent, resulting in differences of up to 49\% relative performance. 

\end{hide_content}
% \cd{Missing a key point somewhere in the preceding paragraphs. \textbf{Here we are primarily motivated by scenarios where an agent would only have access to an image of the GUI environment, such as, ...}}

\begin{figure}[t]
    \centering
    \includegraphics[width=\linewidth]{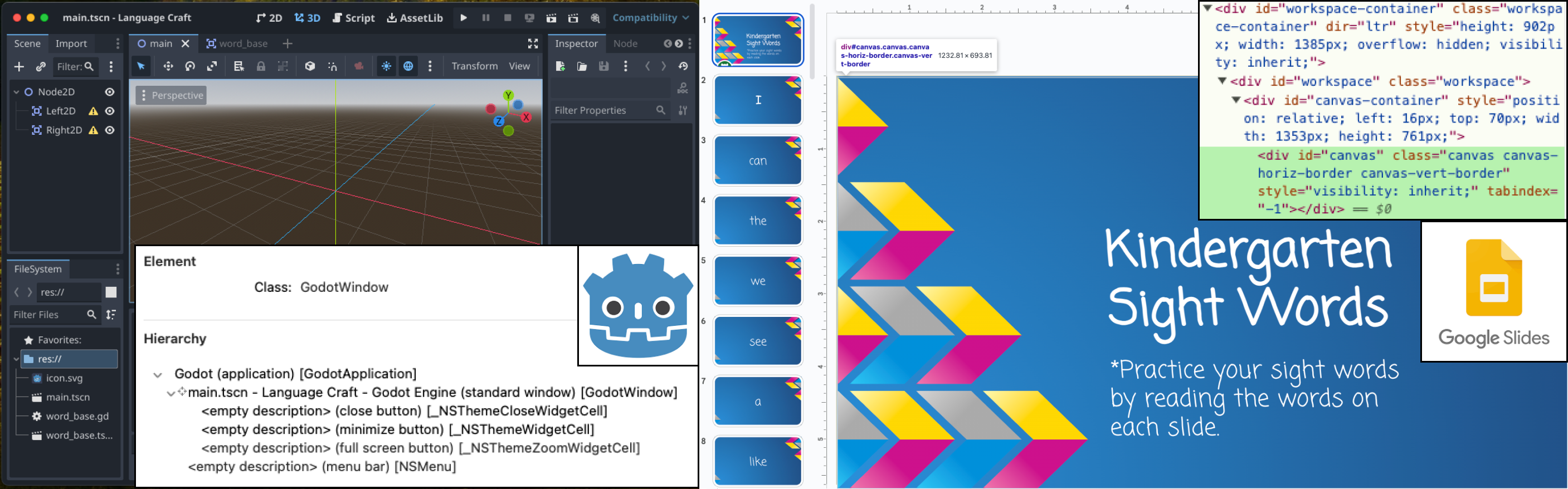}
    \caption{Many software applications lack informative accessibility trees or DOMs. The accessibility tree for a popular game development engine (\textit{Godot}, left) contains only the exit, minimize, and full screen buttons. For a presentation slide (\textit{Google Slides}, right), no interactive elements (e.g. title and subtitle boxes) are present in the DOM. Language model agents rely on this information to navigate applications, and agent performance can accordingly be compromised in scenarios where it is incomplete.}
    \label{fig:bad-examples}
\end{figure}

This can prove problematic as many environments lack obvious methods to both derive and order elements.
For example, many mobile and desktop applications (see Figure~\ref{fig:bad-examples}) do not properly expose interactable elements~\citep{Chen_2020, an-epidemiology-accessibility, zhang2021screen}.
% Test
In such environments, pixels may be the only source of information available.
Previous approaches to deriving interactable elements from pixels either leverage off-the-shelf segmentation models~\citep{yan2023gpt4v, kapoor2024omniact} or build custom models that target accessibility features~\citep{web-ui}.
In our approach, we leverage common crawl~\citep{commoncrawl} to train an object detection model~\citep{ren2016faster} that detects interactable UI elements specifically for agents. 
%To our best understanding, 
To the best of our knowledge, 
the elements detected through these approaches are ordered arbitrarily (e.g. based on confidence scores); 
visually, the ordering is effectively random.
Our experiments indicate that a random ordering consistently results in the lowest performance across multiple scenarios.

%When not explicitly given, an effective ordering of elements is non-trivial to determine due to the hierarchical nature of most environments.
Here we propose and evaluate strategies for deriving effective element orderings in scenarios where a hierarchical ordering based on the GUI design is not explicitly provided by the environment. 
%We propose and evaluate reasonable baselines for deriving orderings. 
% When not explicitly given, an effective ordering of elements is non-trivial to determine.
% Finding a good ordering is surprisingly difficult \cd{is it? this is too subjective. Something more like ``Determining the optimal ordering of elements in a particular scenario with a particular LM is non-trivial.''}; an intuitive raster ordering of elements is insufficient as elements are inherently hierarchical and frequently form local clusters \cd{the ``inherently hierarchical'' claim sanity checks but ``frequently form local clusters'' does not---raster scan should capture some aspects of locality}.
Across multiple agent benchmarks, we find that ordering elements via a 2D-to-1D dimensionality reduction~\citep{van2008visualizing} reliably yields improvements to agent performance relative to other baselines.
We experiment on the VisualWebArena \citep{koh2024visualwebarena} and OmniACT \citep{kapoor2024omniact} benchmarks and achieve new state-of-the-art performance on OmniACT.

Out contributions are as follows.
% \cd{The first sentence of each of these should be concisely convey a reasonably-compelling takeaway.}
\begin{itemize}
    \item We conduct a thorough ablation of VisualWebArena's state representation for agents by including or removing each element attribute individually. Despite advancements in vision language models, we find that a text representation is still necessary for web and desktop agents. We find that element ordering is, perhaps surprisingly, more impactful than any other attribute in the text representation.
    \item We demonstrate that ordering via dimensionality reduction consistently provides performance improvements over random ordering. Additionally, we find that ordering via dimensionality reduction performs better than 
    %raster 
    % CHRIS: I think ``raster'' might not make sense in a vacuum to non CRT enthusiasts
    a simple position-based 
    ordering in most scenarios.
    \item We achieve a new state-of-the-art result on OmniACT, an agent benchmark that considers the scenario of operating on pixels. Our approach more than doubles the expected average task success rate compared to the previous state-of-the-art.
\end{itemize}

%% file: 2_related.tex
\section{Related Work}
\label{sec:related-works}
% While actions may differ slightly (e.g. scrolling vs dragging), at a high-level navigating any of these environments requires an agent to interact with various elements in the GUI.

\paragraph{Agent Benchmarks.} 
World-of-bits provided the first environment for evaluating web GUI navigation using an agent~\citep{world-of-bits}.
Over time, more realistic web~\citep{web-arena, koh2024visualwebarena, web-shop}, desktop~\citep{kapoor2024omniact, xie2024osworld}, and mobile~\citep{rawles2023android} agent benchmarks have been created.
\citet{language-computer-tasks} provided one of the first LM agent approaches, successfully navigating World-of-bits.
However, existing agents are still unable to properly navigate more realistic benchmarks, completing only 15\% of web tasks~\citep{web-arena} and 12\% of desktop tasks.

\paragraph{Agents With Direct Access to Elements.}
Despite work on multimodal agents~\citep{koh2024visualwebarena}, existing techniques in navigating web and desktop environments still rely heavily on ground-truth text representations. 
\citet{web-arena, koh2024visualwebarena} both utilize the accessibility tree and its elements as their state representation. 
\citet{koh2024visualwebarena, he2024webvoyager} consider approaches where interactable elements are also represented in an image via Set-of-Mark~\citep{set-of-mark} bounding boxes and labels.
\citet{xie2024osworld} provides an agent that navigates desktop applications by observing a filtered down version of the accessibility tree.
All of these approaches require access to either a webpage's underlying DOM or an accessibility tree to derive elements;
however, our focus is on environments that only give access to their graphical representations which is significantly more challenging.

\paragraph{Agents With Access to Only a Graphical Representation.} 
There have been several approaches---primarily focused on desktop and mobile environments---to directly navigating a GUI via its pixels.
\citet{kapoor2024omniact} and \citet{yan2023gpt4v} focus on navigating desktop and mobile applications respectively.
Both leverage an off-the-shelf-segmentation model---Segment Anything~\citep{kirillov2023segment}---to find icons in the image. 
These icons are then either represented in text~\citep{kapoor2024omniact} or labeled with Set-of-Mark~\citep{set-of-mark} bounding boxes and labels in the image~\citep{yan2023gpt4v}.
We instead train an object detection model that detects interactable UI elements directly.
While previous UI element detection models are trained to detect accessibility features~\citep{web-ui}, our model is trained specifically to detect interactable elements that would be useful to an agent.
All three of our approaches use Optical Character Recognition (OCR) modules such as EasyOCR~\citep{easyocr} to extract text from pixel information.

\paragraph{Agent Input Ablations.}
While most agent studies include some ablations, few focus on detailed analysis of an agent's input.
To our knowledge, \citet{huq2023whats} is the only other study that directly studies this. 
Their study focuses on broader components to an input prompt such as the selection of few-shot examples used, while we focus on specific element attributes such as element ordering.

%The most common exploration of is mu

% Due to limited context length, it's often infeasible for an LM agent to ingest the entire HTML.
% Various approaches for compressing the HTML have been proposed \citep{mind2web, zheng2024gpt4vision}.
% In particular, the accessibility tree has emerged as a popular state representation across a diverse set of environments\citep{web-arena, koh2024visualwebarena, xie2024osworld}. 
% Unlike other approaches which require modeling approaches, the accessibility tree is simply a filtered version of the HTML representation.
% Despite its popularity, the specific features of the accessibility tree are relatively unexplored.
% We chose the accessibility tree as our default state representation due to its popularity and simplicity and provide an analysis on the important features in an accessibility tree.

%% file: 4_method.tex
\section{Problem Definition}
\label{sec:problem-definition}

We define the environment state $\mathcal{E}$ as a set of elements ${\mathcal{E} = \{e_1, e_2, \ldots, e_n\}}$, 
% We define the environment state $\mathcal{E}$ as the pair of GUI pixels and a set of elements ${\mathcal{E} = \langle \mathcal{P}, \{e_1, e_2, \ldots, e_n\}} \rangle$, 
where each element $e_j$ is a tuple $\langle i_j, C_j, A_j, S_j \rangle$ defined by the following parameters:
\begin{itemize}
\item $i_j \in \{0, 1\}$ denotes the interactability of element $e_j$. An element with $i_j = 1$ is interactable, while an element with $i_j = 0$ is not. %Interactability determines whether an agent can perform actions on the element.
\item $C_j = \{\langle x_1, y_1 \rangle, \langle x_2, y_2 \rangle\}$ is the set of pixel coordinates that form the bounding box around $e_j$. $\langle x_1, y_1 \rangle$ is the top left coordinate and $\langle x_2, y_2 \rangle$ is the bottom right coordinate.
\item $A_j = \{a_1, a_2, \ldots, a_m\}$ is the set of potential actions that can be taken on that element. For example, the potential actions for a search bar might be $\{\texttt{click}, \texttt{type}\}$; a non-interactive text element has the action set $\varnothing$.
\item $S_j$ represents the set of other environment-specific attributes for element $e_j$. These attributes can include image captions, type labels (e.g., Button, Text Field), or accessibility information.
\end{itemize}

While certain environments may provide full access to this environment state, 
%we are focused on virtual environments 
here we are focus on environments where only the pixel information, $\mathcal{P}$, is available.
We then must predict elements from the pixel information to construct a state representation for the agent.
In other words, we must find must find a function $g\colon \mathcal{P} \rightarrow \mathcal{E}$

Elements can be represented in both visual and text modalities. 
% The multi-modal representation allows for a comprehensive understanding of the environment and facilitates various tasks such as element identification, interaction, and manipulation.
For images, the most common approach is to overlay bounding boxes with numeric identifiers around each interactable element \citep{koh2024visualwebarena, he2024webvoyager, yan2023gpt4v} in a manner inspired by Set-of-Mark Prompting \citep{set-of-mark}.
In text, a common approach is to represent each element as a string containing its index, coordinates, and other attributes, such as "[1] [x,y] [Description]".
Because LMs operate on sequential data, elements must be given an ordering; 
in most approaches, this is implicitly defined by the method used to identify the elements.
% For example, the default method to parsing elements in a webpage is a pre-order traversal of the DOM tree \citep{w3c2013selectors}.

% \textbf{Hypothesis}: Our hypothesis is that the ordering of elements \( \{e_1, e_2, \ldots, e_n\} \) in the observation significantly impacts the performance of an agent. 
\paragraph{Ordering.} The ordering is defined as a permutation \( \sigma \) of the indices \( \{1, 2, \ldots, n\} \) where \( \{e_{\sigma(1)}, e_{\sigma(2)}, \ldots, e_{\sigma(n)}\} \) represents a specific sequence of elements.
An ordering function is a function $f\colon \mathcal{E} \rightarrow \sigma$ that takes the environment as an input and yields a specific ordering $\sigma$.
%and maximizes a defined performance criterion. %\( C(\sigma, E) \).
%For agents, this metric is typically the task success rate \citep{web-arena, koh2024visualwebarena, he2024webvoyager}. 

% \textbf{Challenge}: The challenge lies in developing a model or algorithm that can effectively predict or derive the optimal ordering \( \sigma^* \), considering the inherent complexities and variabilities in the element properties and environmental context.

% We construct our state representation based on an increasingly popular representation we describe as \setofactions.
% Inspired by Set-of-Mark Prompting\citep{set-of-mark}, each interactable elements in an application is marked by bounding boxes and alphanumeric labels.
% % While the exact format varies between approaches, each marked element is then listed out in text for the agent to observe (Figure \ref{fig:set-of-action}).
% Each marked element is then listed out in text (the exact format varies between approaches) for the agent to observe (Figure \ref{fig:set-of-action}).
% While other information is frequently present in these representations, the key differentiating factor is that the agent can then decide between marked elements as potential action candidates (e.g. CLICK [1] or CLICK $\langle x, y\rangle$).
% %Unlike Set-of-Mark, 
% We format our representation based on VisualWebArena's representation \citep{koh2024visualwebarena};
% however, the method by which \setofactions\ are created varies depending on information available in an environment.

%% file: 4.1_feature_ablation.tex
\section{Which Aspects Of Agent State Representations Are Most Impactful?}
\label{sec:feature-ablation}

% We analyze each feature in the state representation used in VisualWebArena. 
Here we describe a series of ablation experiments designed to examine which aspects of an LM state representation are most impactful to the performance of LM agents. 
In particular, we experiment on the VisualWebArena (VWA)~\citep{koh2024visualwebarena} benchmark and ablate attributes of the state representation of the state-of-the-art agent (proposed in the same paper). 
We pick this representation in particular due to both its strong performance on VWA, as well as its similarity to common practices seen in other agent research~\citep{he2024webvoyager} and open source projects~\citep{vimGPT, GPT4VAct}. 
Our experiments in turn ablate the impact of 
(1)~multimodal (image and text) aspects of the state representation, and
(2)~individual element attributes within the text component alone.
% Because it is representative of common practices across research and open source projects, 
% element attributes are important to a LM agent, we ablate the state representation of the agent described in . We pick this state representation due to its similarities to a range of .

\subsection{VisualWebArena}
VisualWebArena focuses on multimodal tasks in the web and provides a self-hostable environment for language agents to navigate~\citep{koh2024visualwebarena}.
Agents operating on VisualWebArena have full access to the DOM.
The current best approach~\citep{koh2024visualwebarena} utilizes a multimodal representation where elements are parsed in a pre-order traversal of the DOM tree \citep{w3c2013selectors}. Each element $j$ has attributes 
$S_j = \{id, tag, text\}$ where $id$ is a unique numeric identifier for interactable elements and $\varnothing$ otherwise, $tag$ is the HTML tag (e.g. BTN or IMG), and $text$ is the alt text and captions for images and the HTML text otherwise.
In the text representation, an example of an element would be ``[1] [IMG] [alt text, caption]''.
In the image representation, each element is labeled with bounding boxes and numeric labels.

%Similar state representations have been growing in popularity in both research~\citep{koh2024visualwebarena, he2024webvoyager} and public projects~\citep{vimGPT, GPT4VAct}.
%We experiment with the pre-order  to demonstrate the potential for future improvements to ordering in environments without a DOM.
%In section~\ref{sec:feature-ablation} we also experiment on the impact of each feature in this representation. 
The original agent in~\citep{koh2024visualwebarena} only achieves 15\% success rate across all tasks. 
Since our goal is not to improve agent performance on VisualWebArena but rather to understand the importance of attributes in the state representation, 
we examine a subset of tasks to reduce costs.\footnote{A full run of the state-of-the-art agent on VisualWebArena can cost up to \$800 with GPT-4V.}
% and takes multiple days to complete.} 
Specifically, we explore tasks marked as ``easy'' within tasks that the original agent completed successfully. 
Due to variance associated with stochastic LM outputs, our reproduction of these originally-successful tasks yields a success rate around $74.07\% \pm 5.56\%$. 
The exact list of tasks can be found in Appendix~\ref{sec:subset}.
We reuse the action space from the original agent which consists of executing high-level actions (e.g.,~click, hover) on individual elements---see Appendix~\ref{sec:actions} for more details.

\subsection{Ablation Setup and Findings}

% CHRIS: I am working on this ATM Wayne: OK

The agent state representation we explore is multimodal and consists of image and text information. 
%, primarily derived from the rendered webpage and the DOM respectively. 
The image consists of a screenshot of the webpage along with Set-of-Mark annotations, while the text consists of a DOM-ordered list of elements with the attributes outlined above. 
Our ablation protocol consists of removing individual attributes from the image or text representation and measuring task success rate---we say an attribute has high ``impact'' if its removal leads to substantial reduction in task success rate. 
To provide evidence that these findings may be robust across different LM backbones, 
we explore both GPT-4V as used in the original agent,
and Gemini 1.5 Pro. 
Some experiments were not run on GPT-4V due to high associated costs, 
though we found ordering to be consistent between these two LM backbones on all ablations where we ran both. 

In Table~\ref{tab:multimodal_ablation}, we report results ablating aspects of the multimodal representation. 
In Table~\ref{tab:text_ablation}, we report the impact of ablating various attributes in the text representation specifically. 
Across all experiments, we consider the pre-order traversal of the DOM tree as the ground truth element ordering, and define the ``removal'' of ordering information as substituting an ordering $\sigma_{rand}$ picked uniformly at random from all possible permutations. 
We summarize a few key findings from both sets of ablations below. 

\begin{table}[t]
    \centering
    \caption{Ablating the multimodal aspects of state representation in VisualWebArena. \cmark\ indicates ground truth obtained from the HTML. \xmark\ indicates removal of the attribute. We find that removing the text representation can dramatically harm agent performance.}
    \vspace{5pt}
    \begin{tabular}{ccc|cc}
        \toprule
        \multicolumn{3}{c}{Observations} & \multicolumn{2}{c}{Success Rate (\(\uparrow\))} \\
        \cmidrule(lr){1-3} \cmidrule(lr){4-5}  % Use (lr) to trim the line a bit on both ends
        Screenshot & Set-of-Mark & Text Representation & Gemini 1.5 & GPT-4V  \\
        \midrule
        \cmark & \cmark & \cmark & 64.20\% & 74.07\%  \\
        \xmark & \xmark & \cmark & 46.30\% & 38.89\%  \\
        \cmark & \xmark & \cmark & 50.00\% & -        \\
        \cmark & \cmark & \xmark & 3.70\% & 38.89\%  \\
        % \cmark & \cmark & \pmark & -\% & 46.3\% \\
        % \midrule
        % \cmark & \pmark & \pmark & 51.85\% & 31.48\% \\
        \bottomrule
    \end{tabular}
    \label{tab:multimodal_ablation}
\end{table}

\paragraph{Text Representation is Still Necessary.}
\label{sec:text-needed}
While adding a visual representation clearly improves performance, we find that it alone is insufficient even with Set-of-Mark labels.
This contradicts previous findings on agents for mobile applications which found that a screenshot with Set-of-Mark labels achieves similar performance with or without text~\citep{yan2023gpt4v}.
We 
%believe 
speculate 
that this is due to the 
%vast 
substantial 
difference in viewport sizes between mobile and desktop environments.
Specifically, the average mobile device has a viewport size of 360x800 while the average desktop has a viewport size of 1920x1080~\citep{statcounter2024}. 
Additionally, larger viewport sizes have been shown to improve agent performance in desktop environments~\citep{xie2024osworld}.
%From our experience, this is because 
We speculate that this may be because 
current agents almost never understand when to change the screen view (e.g. by scrolling).

\begin{table}[b!]
    \caption{Ablating attributes of the text component of the VisualWebArena state representation. All results include the screenshot with Set-of-Mark bounding boxes and labels. TAG is the HTML tag. CAPTIONS are image captions generated using BLIP-2-T5XL\citep{li2023blip2}. TEXT$_{Alt}$, TEXT$_{Interact}$, and TEXT$_{Static}$ are the alt text, text for interactable elements, and text for non-interactable elements respectively. ORDER is element ordering. \cmark\ indicates ground truth obtained from the HTML. \xmark\ indicates removal of the attribute. \xmark\ Element Ordering indicates a random shuffling of the elements. - denotes experiments that were not run due to cost.}
    \vspace{5pt}
    \centering
    \small
    \begin{tabular}{cccccc|cc}
        \toprule
        \multicolumn{6}{c}{Element Attributes} & \multicolumn{2}{c}{Success Rate ($\uparrow$)}  \\
        \cmidrule(lr){1-6} \cmidrule(lr){7-8}
        TAG & CAPTIONS & TEXT$_{Alt}$ & TEXT$_{Interact}$ & TEXT$_{Static}$ & ORDER & Gemini 1.5 & GPT4-V\\
        \midrule
        \cmark & \cmark & \cmark & \cmark & \cmark & \cmark & 64.03\% & 74.07\% \\
        \xmark & \cmark & \cmark & \cmark & \cmark & \cmark & 61.11\% & 61.11\% \\
        \cmark & \xmark & \cmark & \cmark & \cmark & \cmark & 46.30\%  & - \\
        \cmark & \cmark & \xmark & \cmark & \cmark & \cmark & 68.15\% & 66.67\% \\
        \cmark & \cmark & \cmark & \xmark & \cmark & \cmark & 53.70\%  & - \\
        \cmark & \cmark & \cmark & \cmark & \xmark & \cmark & 57.40\%  & - \\
        \cmark & \cmark & \cmark & \xmark & \xmark & \cmark & 35.18\% & - \\
        \cmark & \cmark & \cmark & \cmark & \cmark & \xmark & 37.04\% & 44.44\% \\
        % \xmark & \xmark & \xmark & \xmark & \xmark & \xmark & \xmark & 3.7\%   & 38.89\% \\
        \bottomrule
    \end{tabular}
    %\pmark\ indicates a predicted version of the feature.
    \label{tab:text_ablation}
\end{table}

\paragraph{Removing Ordering Information Harms Performance More Than Removing Any Other Attribute.}
% \cd{Maybe would make this a bit more precise. ``Removing ordering information harms performance more than removing any other feature.''}
Although most element attributes are important, we find that ordering is the single most important attribute for agent performance. Random ordering results in a similar performance drop to removing all HTML text descriptions.

\paragraph{Captions Impact Performance More Than Alt Text.}
% \cd{This doesn't sanity check. Removing captions harms task success more than removing alt text, but it's unclear if that means they ``provide more information''}
% \comment{hm... but this is definitely true. I guess I don't have enough evidence here?}
Removing captions causes a greater decrease to performance than removing alt text.
From our experience, captions almost always provide more information than alt text. 
In fact, captions frequently include the alt text directly in its description.

%% file: 4.2_experimental_setup.tex
\section{Experimental Setup}
\label{sec:experimental-setup}

%\cd{I feel like we need a better transition here to allude to the high-level question of section 4 heading. Conceptually, somethign like ``For the remainder of this paper, we focus on applying the insights from our state representation ablations on VWA to a distinct and challenging setting: empowering LM agents to operate directly on pixels.''}

For the remainder of this paper, we leverage the insights gained from our state representation ablations on VisualWebArena to tackle a more challenging task: enabling LM agents to act in environments that only expose pixel information.

Most applications are built on top of an underlying hierarchical representation.
For example, a webpage is modeled by the DOM which is hierarchical.
When exposed, this hierarchy can be used to determine a strong element ordering.
% When the object model is properly exposed, element attributes can be directly extracted. 
% Additionally, the hierarchical nature of object models can be used to determine element ordering.
However, the availability and quality of an underlying hierarchical representation can vary greatly between environments.
For example, \citet{Chen_2020} found that 77\% of mobile applications had missing labels and \citet{an-epidemiology-accessibility} found that 53\% of image buttons had missing labels and were incorrectly sized. 
Additionally, \citet{an-epidemiology-accessibility} found that 
%815 out of 9999 
$8\%$ of 
applications lacked interactable element information altogether.
In such scenarios, we may only have access to the application's pixel information.
We continue to experiment with VisualWebArena and also experiment with the OmniACT benchmark as a scenario where we only have access to pixel information. Details on the LM agent backbones used in our experiments can be found in Appendix~\ref{sec:lm-agent-settings}.
% We experiment on two benchmarks: VisualWebArena \citep{koh2024visualwebarena} where we have access to the DOM and OmniACT where we only have access to the application screenshot.

\subsection{OmniACT}

OmniACT provides both web and desktop environments for agents to benchmark on.
OmniACT contains 177 application screenshots overall and 2021 tasks in the test set. 
Agents are tasked with generating pyautogui code that can navigate the application screenshot.
We consider OmniACT as a setting where only a pixel information is available.

To detect UI elements $\{e_1, e_2, \ldots, e_n\}$ when given only pixel information $\mathcal{P}$ we train an object detection model~\citep{ren2016faster} to detect interactable UI elements in the screenshot and use EasyOCR~\citep{easyocr} to extract visible text.
In other words, the function $g\colon \mathcal{P} \rightarrow \mathcal{E}$ is defined by the trained object detection model.
% For each interactable UI element, we add 
We add visible text and captions to each interactable UI element.
We gather a dataset by finding 67,530 interactable elements over 1468 Common Crawl webpages.
We selected our webpages based on top websites from \citet{similarweb2024}.
Despite the domain shift from webpages to desktop applications, we found that our object detection model worked reasonably well on the OmniACT benchmark in the end-to-end agent setting.
We publicly release this model along with our paper.
Training details can be found in the Appendix~\ref{sec:training-details}.

OmniACT provides partial human annotations for each screenshot; multiple, but not all, interactable UI elements are annotated with bounding boxes. 
The original intent of these bounding boxes is for evaluation only.
As a result, there are significantly less UI elements annotated compared to possible UI elements in the application.
We experiment with elements derived from these human annotated bounding boxes to understand a) impacts to ordering performance in an easier setting and b) the potential performance that can be gained by improving UI element detection.

\subsection{Metrics}

Our primary metric is task success rate which is the standard for agent evaluations~\citep{koh2024visualwebarena, web-arena, he2024webvoyager}.
VisualWebArena provides an evaluation framework for task success.
Task success criteria include achieving an expected final webpage state or receiving a desired response from the language agent.
OmniACT does not provide task success rate directly, instead introducing the sequence score and action score metrics.
Sequence score evaluates if the output contains the correct high level action (e.g. \texttt{click} or \texttt{type}), but does not check if the action element or parameter (e.g. \texttt{click [1]} or \texttt{type [parameter]}) are correct.
Action score evaluates if the output contains both the correct high level action and the correct element or parameter.
Thus, for OmniACT we focus our evaluations on the action score as it is more similar to task success rate.

\subsection{Ordering Methods}

% \cd{I think our ablations should come first before we get into the nitty-gritty about ordering?}

In addition to random ordering, we experiment with two different ordering methods.

\paragraph{Random.} We pick the ordering $\sigma_{rand}$ uniformly at random from all possible permutations. This provides a baseline performance. % but is similar been used by agents in the past \citep{yan2023gpt4v}.
\paragraph{Raster.} Elements $\{e_1, e_2, \ldots, e_n\}$ are ordered in a left-to-right raster scan. We define a raster scan as an ordering $\sigma_{raster}$ where $i < j$ iff $\lfloor \frac{y_i}{8} \rfloor < \lfloor \frac{y_j}{8} \rfloor$ and $x_i < x_j$. We chose to discretize the scan to prevent jumps in ordering from minor pixel variations.
%The y-coordinate space is discritized by a factor of 8 to prevent jumps in ordering from minor pixel variations. 
This method mimics the natural way English speakers read text and images from left-to-right and top-to-bottom.

\paragraph{t-SNE.} We apply dimensionality reduction techniques to better capture the spatial relationships. Using t-SNE~\citep{van2008visualizing}, we reduce the dimensionality with the function $g\colon \langle x, y\rangle \rightarrow z$. 
The set of values $Z = \{z_1, z_2, \ldots, z_m\}$ is used to determine the ordering $\sigma_{tsne}$. We use the scikit-learn~\citep{scikit-learn} implementation of t-SNE and keep the default parameters.

Our intuition for choosing t-SNE stems from visually and qualitatively inspecting our agent trajectories. We observed that agents often use adjacently ordered elements as context clues to determine the correct action. Furthermore, agents have difficulty reasoning about functionally associated elements that are separated from each other in the ordering. t-SNE generates an effective ordering as elements that are visually close together (i.e. nearby in 2D pixel space) are often functionally associated with each other as well. When reducing dimensions, t-SNE retains local structure which increases the odds that functionally associated elements are adjacent in the induced 1D ordering.

\subsection{Action Space}

%\cd{Personal pref but I don't like the inline table. Could we side-by-side this with table 4 and put them both at the top?}

\noindent
\begin{table}[h!]
    \begin{minipage}{0.425\textwidth}
        % Text to the left of the table
        We use the same high-level action space as described in OmniACT.
        Unlike OmniACT, we do not have the model directly output pyautogui code.
        Instead, we use high level actions that map to pyautogui code. 
        For example, for element $e_1$ defined by $\langle 1, \{\langle 50, 50 \rangle, \langle 100, 100 \rangle\}, \{\texttt{click}\}, \varnothing \rangle$ 
        the model would output \texttt{click [1]} which would be converted to \texttt{pyautogui.click(75, 75)}.
        This prevents the model from having to reason about pixel coordinates directly.
        Each element's unique identifier reflects the position of the element in the ordering.
        The full action space is in Table \ref{tab:omniact_actions}.
    \end{minipage}%
    \hfill
    \begin{minipage}{0.55\textwidth}
        \caption{The set of possible actions in OmniACT.}
        \vspace{5pt}
        \centering
        \scriptsize 
        \begin{tabular}{ll}
            \toprule
            \textbf{Action} & \textbf{Description} \\
            \midrule
            \texttt{Click} & Perform a single click on an element. \\
            \texttt{Double Click} & Perform a double click on an element. \\
            \texttt{Right Click} & Perform a right click on an element. \\
            \texttt{Move/Hover} & Move the cursor over an element. \\
            \texttt{Drag} & Click and drag an element to a new position. \\
            \texttt{Scroll} & Scroll up or down the page. \\
            \texttt{Horizontal Scroll} & Scroll left or right on the page. \\
            \texttt{Press} & Press a key on the keyboard. \\
            \texttt{Keyboard Hotkey} & Use a keyboard shortcut or hotkey. \\
            \texttt{Write} & Type text using the keyboard. \\
            \bottomrule
        \end{tabular}
        \label{tab:omniact_actions}
    \end{minipage}
\end{table}

%% file: 5_results.tex
\section{Results}
\label{sec:results}

Our main findings on the impact of ordering are in Table \ref{tab:ordering_methods}. 
%Our ablation into the various attributes in the state representation are in Tables \ref{tab:text_ablation} and \ref{tab:multimodal_ablation}. 
We utilize our various findings to improve upon OmniACT; our experiments against their baseline are in Table \ref{tab:omniact_performance}.

\begin{table}[t!]
\centering
\caption[]{Performance of different ordering methods across various models and information scenarios. The baseline approach for VisualWebArena is the same as their paper. Human annotations are from OmniACT's annotation files. The Faster-RCNN model is trained to detect interactable UI elements from CommonCrawl webpages. VisualWebArena is evaluated on success rate. OmniACT is evaluated on unweighted action score (i.e. each task is weighted equally). We use Llama3-70B for VisualWebArena and Llama3-8B for OmniACT due to its large size\footnotemark. GPT-4v is evaluated on a 100 random tasks for OmniACT; the exact list is in Appendix~\ref{sec:subset}. Gemini-1.5 and GPT-4v are multimodal while Llama3 is text only.}
\vspace{5pt}
\small
\begin{tabular}{@{}lllccc@{}}
\toprule
\multicolumn{3}{c}{Experimental Settings} & \multicolumn{3}{c}{Success Rate} \\
\cmidrule(lr){1-3} \cmidrule(lr){4-6}
Element Source & Benchmark    & Ordering Method & Gemini-1.5 ($\uparrow$) & GPT-4v ($\uparrow$) & Llama3 ($\uparrow$) \\
\midrule
Ground Truth (DOM) & VWA      & Pre-order  & 64.03\%       & 74.07\%             & 27.79\%    \\
&                             & Random    & 37.04\%       & 37.04\%             & 20.37\%                  \\
&                             & Raster    & 38.88\%       & 53.70\%             & 29.63\%                  \\
&                             & t-SNE     & 44.44\%       & 61.11\%             & 24.07\%                  \\
\midrule
Human         & OmniACT       & Random    & 57.29\%       & 61.52\%$^*$         & 28.67\%            \\
Annotations   &               & Raster    & 61.04\%       & 65.88\%$^*$         & 33.65\%                  \\
&                             & t-SNE     & 59.17\%       & 62.11\%$^*$         & 31.99\%   \\
\midrule
Detected      & OmniACT       & Random    & 39.59\%       & 44.63\%$^*$          & 18.88\%           \\
(Faster-RCNN) &               & Raster    & 45.21\%       & 47.38\%$^*$          & 21.58\%                  \\
&                             & t-SNE     & 47.16\%       & 49.18\%$^*$          & 24.61\%   \\
\bottomrule
\end{tabular}
\label{tab:ordering_methods}
\end{table}

\subsection{Impact of Ordering}

\paragraph{Ordering Consistently Impacts Performance.}
Ordering 
%\cd{\emph{Interpretable} orderings? Random is also an ordering...} \comment{And random has an impact! :D} 
has a significant impact to performance across all of our experiments.
Random ordering decreases performance in VisualWebArena by 50\% and 42\% relative performance for GPT-4v and Gemini-1.5 respectively.
In all experiments, random ordering decreases performance over a proper ordering method.

\paragraph{t-SNE Best For Larger Models And More Challenging Tasks.}
% Both raster and t-SNE ordering consistently improved performance over random ordering.
Navigating by using detected elements is a harder task than navigating by using human annotated bounding boxes;
not only are there more elements---on average double the amount---there is the possibility that the correct element is missing from the detected elements.
We see that when elements are derived from the DOM and our UI detection model, t-SNE ordering generally outperforms raster ordering.
Additionally, more powerful models see an increased benefit from t-SNE ordering with Gemini-1.5 and GPT-4v seeing larger improvements than LLama3.
%These are more challenging settings than when given ground truth human annotations due to the limited amount of human annotations per task.

\paragraph{Raster Ordering Performs Best With Human Annotations.}
Raster ordering performs the best with human annotated elements.
Unfortunately, these annotations are fewer and partially guaranteed to contain important information. 
Additionally, high quality human annotations are difficult to scale across applications.

\begin{hide_content}
\textbf{Ordering Impact Increases with Element Count}\quad
As element count increases, we observe that the impact of ordering increases as well. This is consistent across all datasets, although VisualWebArena shows the greatest discrepancy. Our results on Gemini are in Figure \ref{fig:element_count} and the remainder are in the Appendix.
\end{hide_content}

% Second, human annotations provide significantly fewer elements when compared to a UI element detection model. The median number of human annotated elements is 19 per screenshot; the median  number of detected UI elements is almost double that at 35 elements per screenshot.

\footnotetext{We use the Groq API for our Llama3 models.}

\begin{hide_content}
\begin{figure}[h!]
    \centering
    \includegraphics[width=\linewidth]{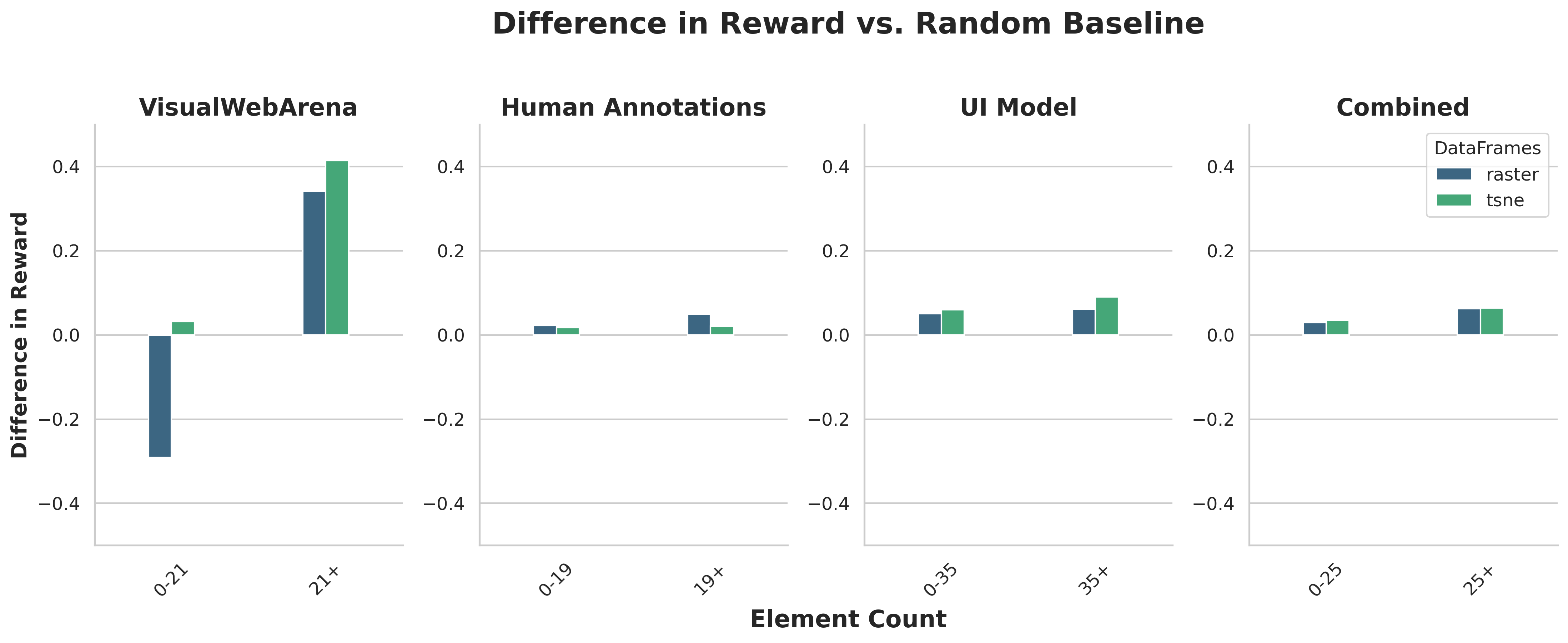}
    \caption{Differences in reward from random ordering. Element counts are divided based on the median. Across all datasets and element sources, the impact of a proper ordering method increases with element count.}
    \label{fig:element_count}
\end{figure}
\end{hide_content}

\begin{hide_content}
Findings:
\begin{enumerate}
    \item Tag and Alt Text are more important for better models. 
    \item Captions, text for interactive elements, and text for static elements are important.
    \item Ordering of elements is surprisingly important.
\end{enumerate}
\comment{I feel like I need another model here. Maybe an OSS one since GPT-4V is expensive.}
\end{hide_content}

\begin{hide_content}
Findings:
\begin{enumerate}
    \item The accessibility tree is the most important.
    \item Set-of-Mark labels are important. It seems as though Set-of-Mark is what enables GUI understanding. \comment{I did an experiment with a white background and bounding boxes. I noticed almost no difference from having the actual screenshot + set-of-marks. Should I write about that?}
    \item Our approach is not able to reach the same performance as with full Accessibility Tree info.
\end{enumerate}
\end{hide_content}

\subsection{State-of-the-Art Performance on OmniACT}
\label{sec:sota}

We achieve a new state-of-the-art performance on Omniact.
Due to cost, we look to our previous experiments to pick the best combination of features for our approach.
We observe that multimodal representations are still helpful.
We find that t-SNE ordering improves performance the best in OmniACT when elements are detected by our model.
\citet{koh2024visualwebarena} states that high level actions are easier for a LM to reason with. 

We analyze the differences between our best approach and OmniACT's baseline.
These are as follows.
\begin{itemize}
    \item \textbf{Element Source:} 
    OmniACT obtains elements by searching for icons with Segment Anything \citep{kirillov2023segment} and text with EasyOCR. Unfortunately, there are no shared artifacts for their icon detection system.
    We obtain UI elements through an object detection model and text with EasyOCR~\citep{easyocr}. 
    \item \textbf{Ordering:} 
    It is unclear how elements are ordered in OmniACT. Considering how most approaches don't pay specific attention to ordering, we assume the ordering in OmniACT is effectively random.
    We order our elements using our t-SNE ordering. 
    \item \textbf{Action Space:} 
    OmniACT directly outputs pyautogui code as their actions. 
    We consider a higher level action space that maps to pyautogui code.
    \item \textbf{Intractability} OmniACT lists out each element, but does not indicate which element is interactable. We specify whether elements are interactable or not.
    \item \textbf{Multimodal Representation:} OmniACT evaluates their full test set using a text only representation\footnote{OmniACT evaluates the impact of adding a visual representation on smaller subset; however, this subset is not shared and varies empirically from the full test set.}. We experiment with a multimodal representation.
\end{itemize}

% Due to cost, we only ran the experiments we though
% We experiment with different different action spaces, orderings, and representation modalities. Our results are in Table \ref{tab:omniact_ablations}. We find that the best result is the combination of a pyautogui code action space, t-SNE ordering, and a text only representation. This is surprising as a multimodal representation typically improves results. \comment{But then the story sort of falls apart here because we didn't run this result.}
We apply our findings and achieve more than one-fold increase over the existing best action score. Our results can be found in Table~\ref{tab:omniact_performance}.

\begin{hide_content}
\begin{table}[h]
    \centering
    \begin{tabular}{llc|cc}
        \toprule
        Action Space & Ordering & Screenshot & Action Score ($\uparrow$) & Sequence Score ($\uparrow$)\\
        \midrule
        Code & Random &\xmark & 15.89  & 21.14 \\
        & Random & \cmark & 25.16  & 33.12 \\
        & t-SNE  & \xmark & 27.80 & 35.43 \\
        & t-SNE  & \cmark & 24.64 & 29.51 \\
        \midrule
        Actions & Random & \xmark  & 13.83 & 18.02 \\
        & Random &\cmark & 19.86  & 25.55 \\
        % & Random & \cmark  & 13.83 & 18.02 \\
        & t-SNE & \xmark & 15.03 & 20.86 \\
        & t-SNE & \cmark & 25.04 & 29.39 \\
        \bottomrule
    \end{tabular}
    \vspace{3pt}
    \caption{GPT-4 with various action spaces and ordering methods. Actions indicates actions such as CLICK [1] which are converted to pyautogui code. Code indicates an action space where pyautogui code is directly generated. Experiments are run on a random sample of 100 tasks. t-SNE ordering provides a significant improvement to performance over random when directly generating pyautogui code.}
    \label{tab:omniact_ablations}
\end{table}
\end{hide_content}

\begin{table}[t]
    \centering
    \caption{\cmark\ and \xmark\ indicates whether the feature is available when building a representation for the model. Ours indicates a high level actions such as \texttt{click [1]}. Code indicates that pyautogui code is directly generated. \textbf{bold} is best and \textit{italics} are second best. While sequence score only checks for the correct high level action (e.g. \texttt{click}), action score checks for both the correct action and the correct element or parameter (e.g. \texttt{click [1]}). Thus, action score is the most equivalent to task success rate. * is as reported in~\citet{kapoor2024omniact}}
    \vspace{5pt}
    \begin{tabular}{lllc|cc}
        \toprule
        Model & State & Actions & Screenshot & Action Score ($\uparrow$) & Sequence Score ($\uparrow$)\\
        \midrule
        % Ours & llama8b & Code & \cmark & \xmark & 8.25* & 5.26* \\
        % Ours & Gemini 1.5 & Code & \cmark & \cmark & 28.43 & 23.8 \\
        % Ours & llama8b & Actions & \cmark & \xmark & 23.19 & 17.95 \\
        % Ours & Gemini 1.5 & Actions & \cmark & \cmark & 28.82 & 25.01 \\
        % Ours & Gemini 1.5 & Actions & \cmark & \xmark & 32.07 & 27.00 \\
        % % omniact_gemini_single_best_gt_bbox_all_20240509_232928
        % \midrule
        GPT-4       & OmniACT  & Code & \xmark           & 11.60*  & \textbf{32.75}* \\
        Gemini 1.5  & Ours     & Code & \xmark   & 16.53 & 21.67 \\
        Gemini 1.5  & Ours     & High Level & \xmark   & 22.29 & 29.42 \\
        Gemini 1.5  & Ours     & High Level & \cmark   & \textit{22.86} & 28.91 \\
        Llama8b     & Ours     & High Level    & \xmark      & 18.64 & 26.22 \\
        GPT-4v      & Ours     & High Level     & \cmark       & \textbf{23.34} & \textit{30.47} \\
        % Ours & GPT-4  & Code     & \cmark       & - & - \\
        % Ours & Gemini 1.5 & Code & \pmark & \cmark & - & - \\
        % Ours (no OCR in interaction text)& llama8b & Actions & \cmark & \xmark & 20.37 & 15.48 \\
        \bottomrule
    \end{tabular}
    \vspace{1pt}
    \label{tab:omniact_performance}
\end{table}

%% file: 6_future_work.tex
\section{Future Work}
\label{sec:future-work}

\paragraph{Further Improvements to Ordering}
% PLACEHOLDER
% Hierarchy
% Better clustering or graph traversal
We provided two simple methods to apply ordering when a default ordering is not given.
However, both approaches still fall short when compared to the hierarchical ordering derived from the DOM.
We hope that future research can introduce more sophisticated methods to find ordering with only pixel information.
% We hope that future research can continue to close this performance gap, even without access to a DOM.

\paragraph{Image Only Ordering}
We focused on the impact of element ordering in a text representation (although element labels re-ordered accordingly in the visual representation as well).
The impact of element ordering may or may not generalize to an image only representation.
Unfortunately, our results indicated that a visual representation alone was insufficient for web and desktop environments which prevents us from conducting this experiment.
However, \citet{yan2023gpt4v} found that a visual representation in mobile environments was able to achieve comparable performance with and without a text representation.
In the future, we hope to experiment with various ordering methods on an image only representation.

\paragraph{Expanded Scenarios and Benchmarks}
In this paper, we explored two benchmarks---VisualWebArena and OmniACT---as web and desktop scenarios.
In the future, we hope to explore other benchmarks and settings with our approach.
For example, ~\citet{xie2024osworld} uses the OS level accessibility tree for desktop agent navigation.
We hope to compare our approach against theirs and believe that combining both approaches may lead to further improvements.
Additionally, mobile environments often have only pixel-level information~\citep{Chen_2020, an-epidemiology-accessibility}; we hope to apply our approach to a mobile benchmark such as \citet{rawles2023android}.

% It is crucial to address these issues before deploying LM agents in real-world settings.
% To address these concerns, it is crucial to establish strong safeguards, transparent data management practices, and clear ethical guidelines to ensure that the development and deployment of language model agents are conducted responsibly.
% This prevents us from experimenting if ordering affects 
% We hope to experiment on a mobile 
% Even in the web scenario, there can be improvements. Cite visual web arena, omniact, etc.

% \paragraph{More Realistic Environments}
% While VisualWebArena is the most realistic web benchmark available, it is still relatively clean and structured when compared to real websites.
% We did not fully explore the space of possible scenarios and environments.
% VisualWebArena webpages are much cleaner than many webpages that agents might need to operate on. For example, it contains no advertisements, etc. It would be nice to test on real webpages.
% OmniACT does not have multiple steps, hope to test on an environment like OS World.

%% file: 7_conclusion.tex
\section{Conclusion}
\label{sec:conclusion}

We conducted thorough ablations to show that element ordering has a significant impact on the performance of agents.
We provided a method of ordering elements through dimensionality reduction and showed that it performed best in realistic environments.
We trained a UI element detection model on Common Crawl data and publicly share the model.
We demonstrated an end-to-end method which allows a LM agent to act on environments that provide only pixel information.
Using this method, we were able to achieve a new state-of-the-art performance on OmniACT.

%% file: 10_appendix.tex
\newpage
\section{Appendix}

\subsection{Faster-RCNN Training Details}
\label{sec:training-details}
We trained our model using the detectron2~\cite{wu2019detectron2} implementation of faster-rcnn. We did not change much from the default implementation and recognize that there are significant improvements that could be made to the model. 

\begin{table}[h]
\centering
\begin{tabular}{ll}
    \toprule
    \textbf{Hyperparameter} & \textbf{Value} \\
    \midrule
    Base Learning Rate & 0.00025 \\
    Number of Classes & 1 \\
    Iterations & 200000 \\
    Optimizer & SGD \\
    Backbone & Resnet-50 (ImageNet Pretrained) \\
    ResNet Depth & 50 \\
    Images per Batch & 16 \\
    Objects per Image & 128 \\
    Devices & 8 \\
    \bottomrule
\end{tabular}
\caption{Key hyperparameters for the Faster-RCNN model.}
\label{tab:hyperparameters}
\end{table}

We share the remaining hyperparmeters in a config file.
We also share the model artifacts and dataset.

\subsection{VisualWebArena Agent Action Space}
\label{sec:actions}

% \todo{Do we need to save space? I also think this whole subsection could go in the appendix and be replaced with one sentence (uncomment in the paragraph above)}
We use the same action space as VisualWebArena for all VisualWebArena experiments.

% \vspace{-30pt}
\noindent
\begin{table}[h!]
    \begin{minipage}{0.425\textwidth}
        % Text to the left of the table
        We use the same action space as described in VisualWebArena.
        VisualWebArena uses high level actions that act directly on elements rather than pixel coordinates.
        Interactable elements possess a unique \texttt{id} identifier while non-interactable elements do not.
        The \texttt{id} identifier reflects the position of the element in the ordering.
        The full action space is in Table \ref{tab:vwa_actions}.
    \end{minipage}%
    \hfill
    \begin{minipage}{0.55\textwidth}
        \centering
        \scriptsize 
        
        \begin{tabular}{ll}
            \toprule
            \textbf{Action} & \textbf{Description} \\
            \midrule
            \texttt{click [id]} & Click on element \texttt{id}. \\
            \texttt{hover [id]} & Hover on element \texttt{id}. \\
            \texttt{type [id] [text]} & Type text on element \texttt{id}. \\
            \texttt{press [key\_comb]} & Press a key combination. \\
            \texttt{new\_tab} & Open a new tab. \\
            \texttt{tab\_focus [index]} & Focus on the i-th tab. \\
            \texttt{tab\_close} & Close current tab. \\
            \texttt{goto [url]} & Open \texttt{url}. \\
            \texttt{go\_back} & Click the back button. \\
            \texttt{go\_forward} & Click the forward button. \\
            \texttt{scroll [up|down]} & Scroll up or down the page. \\
            \texttt{stop [answer]} & End the task with an optional output. \\
            \bottomrule
        \end{tabular}
        \vspace{1pt}
        \caption{The set of possible actions in VisualWebArena.}
        \label{tab:vwa_actions}
    \end{minipage}
\end{table}
\vspace{-25pt}

\subsection{LM Agent Hyperparameters and Settings}
\label{sec:lm-agent-settings}

\begin{table}[h!]
    \begin{minipage}{0.425\textwidth}
    We set our temperature, top-p, and input token limits based on existing works~\cite{web-arena,koh2024visualwebarena,kapoor2024omniact}. 
    Prompts for all three backbones contain few-shot examples and use chain-of-thought prompting~\cite{wei2023chainofthought} as described in~\citet{koh2024visualwebarena}.
    In GPT-4v and Llama3 each example is a different message;
    Gemini-1.5's context length allowed us to input all examples as a single prompt.
    We detail our LM Agent hyperparameters in Table~\ref{tab:lm_settings}
    \end{minipage}%
    \hfill
    \begin{minipage}{0.55\textwidth}
        \centering
        \small
        \begin{tabular}{l|ccc}
            \toprule
            \textbf{Setting} & \multicolumn{3}{c}{\textbf{Language Model Backbone}} \\
            \cmidrule(lr){2-4}
            & \textbf{GPT-4v} & \textbf{Gemini-1.5} & \textbf{Llama3} \\
            \midrule
            Input Token Limit & 3840 & 900000 & 3840 \\
            Temperature & 1.0 & 1.0 & 1.0 \\
            Top-p & 0.9 & 0.9 & 0.9 \\
            \bottomrule
        \end{tabular}
        \caption{Settings for different LM agent backbones.}
        \label{tab:lm_settings}
    \end{minipage}
\end{table}

\

\subsection{VisualWebArena and OmniACT Subset}
\label{sec:subset}

We experimented with subsets of VisualWebArena and OmniACT to save on costs. We list them here for reproducibility.

For all VisualWebArena experiments we used the following:

\noindent
\texttt{[13, 15, 50, 129, 164, 167, 0, 77, 86, 89, 98, 99, 100, 101, 105, 130, 131, 142, 143, 146, 150, 189, 16, 29, 37, 38, 39, 47, 49, 52, 53, 56, 60, 61, 62, 69, 73, 76, 77, 81, 148, 173, 193, 196, 201, 212, 216, 231, 273, 314, 315, 322, 445]}

For GPT-4v ordering ablations on OmniACT we used the following:

\noindent
\texttt{[4, 58, 115, 147, 156, 162, 165, 178, 179, 194, 204, 218, 235, 240, 248, 297, 353, 374, 391, 392, 395, 404, 409, 419, 434, 462, 487, 492, 517, 533, 556, 573, 598, 658, 667, 673, 678, 719, 795, 827, 896, 910, 944, 961, 975, 1018, 1025, 1038, 1084, 1093, 1101, 1103, 1128, 1130, 1138, 1142, 1147, 1181, 1192, 1219, 1252, 1284, 1291, 1353, 1427, 1442, 1448, 1514, 1521, 1538, 1580, 1590, 1594, 1600, 1606, 1622, 1636, 1641, 1665, 1684, 1694, 1696, 1710, 1711, 1719, 1726, 1731, 1740, 1743, 1845, 1877, 1883, 1918, 1924, 1951, 1960, 1993, 1994, 1997, 2011]}

\subsection{Prompt}

\begin{lstlisting}[backgroundcolor=\color{codebg}, basicstyle=\ttfamily\small, breaklines=true, backgroundcolor=\color{lightgray}, frame=single, caption={Language Model Prompt}]
You are an autonomous intelligent agent tasked with navigating desktop and web applications. You will be given tasks that can be accomplished by various actions that will be mapped to pyautogui code.

Here's the information you'll have:
The user's objective: This is the task you're trying to complete.
The current desktop screenshot: This is a screenshot of the desktop or webpage, with each interactable element assigned a unique numerical id. Each bounding box and its respective id shares the same color.
The observation, which lists the IDs of all interactable elements on the current screenshot with their text content if any, in the format [id] [tagType] [text content]. tagType is the type of the element. text content is the text content of the element. For example, [1234] [BUTTON] ['Add to Cart'] means that there is a button with id 1234 and text content 'Add to Cart' on the current web page. [] [StaticText] [text] means that the element is of some text that is not interactable.

The actions you can perform fall into two categories:

Mouse Actions:
click [id]: This action clicks on an element with a specific id.
double_click [id]: This action double clicks on an element with a specific id.
right_click [id]: This action right clicks on an element with a specific id.
hover [id]: Hover over an element with id.

Keyboard Actions:
type [content]: Use this to type content. Be sure to use other commands to click before or press enter after if necessary.
press [key_comb]: Simulates the pressing of a key combination on the keyboard (e.g., enter).
hotkey [key1] [key2]: Simulates the pressing of a multiple key combinations on the keyboard. For example, hotkey [Ctrl] [Alt] [Delete] will press Ctrl+Alt+Delete.

To be successful, it is very important to follow the following rules:
1. You should only issue actions that are valid given the current observation. Everything is possible. You MUST issue actions.
2. You can issue a sequence of actions separated by newlines.
3. You should follow the examples from past messages to reason step by step and then issue the next actions.
4. You should start every answer with "Let's think step-by-step"
5. Generate the actions in the correct format. Start with a "In summary, the next actions I will perform are" phrase, followed by the actions inside ```. Each action should be split by a newline. There should be no text inside `` except for the actions. For example, "In summary, the actions I will perform are ```click [1234] type [sample text] press [enter]```".

Here are a few examples:
Example 1:

{Example 1}

Example 2:

{Example 2}

Example 3:

{Example 3}

Those were the examples. Now make a prediction given the observation.

OBSERVATION:

{Observation}

\end{lstlisting}

\begin{hide_content}
\subsection{Full Omniact Experiments}

\begin{table}[ht]
    \centering
    \begin{tabular}{lc|cc}
        \toprule
        Approach & Model & Sequence Score & Action Score \\
        \midrule
        Omniact & GPT-4V & 39.43 & 20.76 \\
        Ours (Actions) w/ GT BBox & Gemini 1.5 & 28.23 & 21.67 \\
        Ours (Actions) w/ GT BBox & GPT-4V & 32.00 & 24.81 \\
        Ours (Actions) w/ GT BBox and Short & Gemini 1.5 & 22.8 & 17.29 \\
        Ours (Pyautogui) w/ GT BBox & Gemini 1.5 & 28.28 & 22.38 \\
        Ours (Actions) w/ GT BBox and filtering & Gemini 1.5 & 25.75 & 19.10 \\
        Ours (Actions) w/ GT BBox and text interaction & Gemini 1.5 & 26.35 & 20.59 \\
        \bottomrule
    \end{tabular}
    \caption{Comparison of different approaches and models on performance metrics. 500 samples. Random Seed 777}
    \label{tab:model_performance}
\end{table}

\begin{table}[ht]
    \centering
    \begin{tabular}{lc|cc}
        \toprule
        Approach & Model & Sequence Score & Action Score \\
        \midrule
        Omniact & GPT-4V & 39.43 & 20.76 \\
        Ours (Actions) w/ GT BBox & Gemini 1.5 & 30.45 & 22.75 \\
        Ours (Actions) w/ GT BBox and Emphasize Ordering & Gemini 1.5 & 33.53 & 25.16 \\
        Ours (Actions) w/ GT BBox, Emphasize Ordering, OCR + Captions & Gemini 1.5 & 31.92 & 25.34 \\
        Ours (Actions) w/ VWA BBox, Emphasize Ordering, OCR + Captions & Gemini 1.5 & 27.41 & 18.55 \\
        \bottomrule
    \end{tabular}
    \caption{Comparison of different approaches and models on performance metrics. 500 samples. Random seed 77.}
    \label{tab:model_performance}
\end{table}
\end{hide_content}

% \subsection{Segment Anything and WebUI}

\begin{hide_content}
\subsection{Table with Scaled Values}

\begin{table}[ht!]
\centering
\begin{tabular}{@{}lllccc@{}}
\toprule
Scenario      & Dataset       & Ordering Method & Gemini-1.5 ($\uparrow$) & GPT-4v ($\uparrow$) & Llama3 ($\uparrow$) \\
\midrule
Object        & VWA           & Baseline        & 64.03\%       & 74.07\%    & 27.79\%    \\
Model         &                & Random          & 37.04\%                & 44.44\%             & 20.37\%                  \\
&                              & Raster          & 37.04\%                & 44.44\%             & 20.37\%                  \\
&                              & t-SNE           & 44.44\%       & 61.11\%                   & 24.07\%                  \\
\midrule
Human         & OmniACT       & Random    & 57.29\%(25.31)                & -                   & 28.67\%(17.34)            \\
Annotations   &                & Raster    & -                   & -                   & -                  \\
&                              & t-SNE     & 59.17\%(25.01)       & -                   & 31.99\%(17.95)   \\
\midrule
Screenshot    & OmniACT     & Random    & 39.59\%(20.27)                     & -                   & 18.88\%(16.84)           \\
Only          &              & Raster    & -                     & -                   & -                  \\
&                            & t-SNE     & 47.16\%(22.86)       & -                   & 24.61\%(18.64)   \\
\bottomrule
\end{tabular}
\vspace{1pt}
\caption{Performance of different ordering methods across various models and information scenarios. The baseline approach for VisualWebArena is the same as their paper, the baseline approach for OmniACT* is taken from their annotation files, and the baseline approach for OmniACT is the ordering given by our object detection model. VisualWebArena is evaluated on success rate. OmniACT* and OmniACT are evaluated on action score. We use Llama3-70B for VisualWebArena and Llama3-8B for OmniACT* and OmniACT. %\textbf{Bolded} are best and \textit{italicized} are second best.}
}
\label{tab:ordering-methods-sequence}
\end{table}

\subsection{Sequence Score Values}

\begin{table}[ht]
    \centering
    \begin{tabular}{lcccc|cc}
        \toprule
        Approach & Model & Action Space & Annotations & Screenshot & Sequence Score ($\uparrow$) & Action Score ($\uparrow$)\\
        \midrule
        Ours & llama8b & Code & \cmark & \xmark & 8.25* & 5.26* \\
        Ours & Gemini 1.5 & Code & \cmark & \cmark & 28.43 & 23.8 \\
        Ours & llama8b & Actions & \cmark & \xmark & 23.19 & 17.95 \\
        Ours & Gemini 1.5 & Actions & \cmark & \cmark & 28.82 & 25.01 \\
        Ours & Gemini 1.5 & Actions & \cmark & \xmark & 32.07 & 27.00 \\
        % omniact_gemini_single_best_gt_bbox_all_20240509_232928
        \midrule
        Omniact & GPT-4 & Code & N/A & \xmark & 32.75 & 11.6 \\
        Ours & Llama8b & Actions & \pmark & \xmark & 26.22 & 18.64 \\
        Ours & Gemini 1.5 & Actions & \pmark & \cmark & 28.91 & 22.86 \\
        Ours & Gemini 1.5 & Actions & \pmark & \xmark & - & - \\
        Ours & GPT-4V & Actions & \pmark & \cmark & 30.47 & 23.34 \\
        Ours & GPT-4 & Actions & \pmark & \xmark & - & - \\
        % Ours & Gemini 1.5 & Code & \pmark & \cmark & - & - \\
        % Ours (no OCR in interaction text)& llama8b & Actions & \cmark & \xmark & 20.37 & 15.48 \\
        \bottomrule
    \end{tabular}
    \caption{\cmark\ and \xmark\ indicates whether the feature is available when building a representation for the model. \pmark\ indicates the feature is predicted using our object detection model. Actions indicates actions such as CLICK [1] which are converted to pyautogui code. Code indicates an action space where pyautogui code is directly generated.}
    \label{tab:model_performance}
\end{table}
\end{hide_content}